\documentclass{article}

\usepackage{amsmath}
\usepackage{amsfonts}
\usepackage{hyperref}
\usepackage{graphicx}
\usepackage{caption}
\usepackage{subcaption}

\title{Linking GloVe with \texttt{word2vec}}

\author{Tianze Shi and Zhiyuan Liu\\\texttt{stz11@mails.tsinghua.edu.cn, liuzy@tsinghua.edu.cn}}
\date{November 20, 2014}
\begin{document}

\maketitle
The Global Vectors for word representation (GloVe), introduced by Jeffrey Pennington et al. \cite{glove}\footnote{\url{http://nlp.stanford.edu/projects/glove/}} is reported to be an efficient and effective method for learning vector representations of words. State-of-the-art performance is also provided by skip-gram with negative-sampling~(SGNS) \cite{sgns} implemented in the \texttt{word2vec} tool\footnote{\url{https://code.google.com/p/word2vec/}}.

In this note, we explain the similarities between the training objectives of the two models, and show that the objective of SGNS is similar to the objective of a specialized form of GloVe, though their cost functions are defined differently.

\section{Introduction and Notation}
By representing words as vectors, similarities between words and other valuable features can be calculated directly with vector arithmetics. The goal of word embedding algorithms is to find vectors for the words and their contexts in the corpus to meet some pre-defined criterion (e.g. to predict the surrounding context of a given word), where the contexts are often defined as the words surrounding a given word.

Let the word and context vocabularies be $V_W$ and $V_C$ respectively. For each word $w \in V_W$ and each context $c \in V_C$, the goal is to find a vector $\vec{w} \in \mathbb{R}^d$ and $\vec{c} \in \mathbb{R}^d$, where $d$ denotes the vector dimension. Embeddings of all words in the vocabulary can be combined into a $\|V_W\| \times d$ matrix $W$, with the $i$th row $W_i$ being the embedding of the $i$th word in the vocabulary. Similarly, a $\|V_C\| \times d$ matrix $C$ gathers all the embeddings of the contexts, with $C_j$ representing the embedding of the $j$th context.

Word-context pairs are denoted as $(w, c)$, and $\#(w, c)$ counts all observations of $(w, c)$ from the corpus. We use $\# (w) = \Sigma_{c}{\# (w, c)}$ and $\# (c) = \Sigma_{w}{\# (w, c)}$ to refer to the count of occurrences of a word (context) in all word-context pairs. Either $\Sigma_{c}{\# (c)}$ or $\Sigma_{w}{\# (w)}$ may represent the count of all word-context pairs.

\section{Training Objectives of the Two Models}

\subsection{GloVe}

GloVe explicitly factorizes the word-context co-occurrence matrix. The following equation gives the local cost function of GloVe model.
\begin{equation} \label{eq:J_glove}
l_{G}\left(w_i, c_j\right) =
f\left(\#\left(w_i, c_j\right)\right)\left(W_{i}\cdot C_{j}^T + b_{W_i} + b_{C_j}
- \log{\#\left(w_i, c_j\right)}\right)^2,
\end{equation}
where $b_{W_i}$ and $b_{C_j}$ are the unknown bias terms only relevant to the words and contexts respectively. $f\left(x\right)$ is a weighting function which down-weights rare co-occurrences. The function chosen by Pennington et al. is
\begin{equation}\label{eq:glove_f}
f\left(x\right) =
\begin{cases}
\left(x/x_{\text{max}}\right)^\alpha & x < x_{\text{max}} \\
1 & \text{otherwise}
\end{cases}
\end{equation}

The cost function is minimized by optimizing $W_{i}\cdot C_{j}^T$ to $\log{\#\left(w_i, c_j\right)} - b_{W_i} - b_{C_j}$, an ideal solution
to which is given by
\begin{equation}\label{eq:wc_glove}
W_{i}\cdot C_{j}^T = \log{\#\left(w_i, c_j\right)} - b_{W_i} - b_{C_j}
\end{equation}
for each row in $W$ and $C$.

\subsection{Skip-gram with Negative Sampling (SGNS)}
As shown by Levy and Goldberg~\cite{sgns-fact}, SGNS implicitly factorizes a word-context matrix, whose cells are the shifted point-wise mutual information (PMI). The local objective for a given word-context pair is
\begin{equation}\label{eq:l_sgns}
l_{S}\left(w_i, c_j\right) =
\#\left(w_i, c_j\right)\log{\sigma\left(W_{i}\cdot C_{j}^T\right)}
+ k \cdot \#\left(w_{i}\right) \cdot \frac{\#\left(c_{j}\right)}{\Sigma_{w}{\#(w)}}
\log{\sigma\left(-W_{i}\cdot C_{j}^T\right)},
\end{equation}
where $\sigma\left(x\right) = \frac{1}{1 + e^{-x}}$ and $k$ is the number of ``negative'' samples.

To optimize the objective, we find its partial derivative with respect to $x := W_{i}\cdot C_{j}^T$ and compare it to zero: 
\begin{equation}\label{eq:pd_sgns}
\frac{\partial l_{S}}{\partial x} = 
\#\left(w_i, c_j\right)\sigma\left(-x\right)
- k \cdot \#\left(w_{i}\right) \cdot \frac{\#\left(c_{j}\right)}{\Sigma_{w}{\#(w)}}
\sigma\left(x\right) = 0. 
\end{equation}
This equation is solved by 
\begin{equation}\label{eq:wc_sgns}
\begin{split}
W_{i}\cdot C_{j}^T &= PMI\left(w_i, c_j\right) - \log{k} \\
&= \log{\#(w_i, c_j)} - \log{\#(w_i)} - \log{\#(c_j)} + \log{\Sigma_{w}{\#(w)}} - \log{k}.
\end{split}
\end{equation}

\subsection{Similarities between the Two Objectives}
By comparing Equation \ref{eq:wc_glove} and \ref{eq:wc_sgns}, we find that they show somewhat similar forms. The $\log{\#(w_i)}$ and $\log{\#(c_j)}$ terms in Equation \ref{eq:wc_sgns} can be absorbed into the bias terms $b_{W_i}$ and $b_{C_j}$ respectively, and the $\log{\Sigma_{w}{\#(w)}} - \log{k}$ term is independent of $i$ and $j$ and can be viewed as a global bias term, which may be divided into the word and context bias terms.

The bias terms in the GloVe objective function are unknown and are to be determined by matrix factorization algorithms. They may or may not converge to the values given in the SGNS objective function. From this perspective, the GloVe model is more general and has a wider domain for optimization.

\subsection{Differences between the Two Objectives}
The GloVe model and the SGNS model are different in the following two aspects.

First, they define different cost functions though they share similar objectives, which may affect the performance when the vector dimensionality is not high enough.

They are also different in weighting strategies. With a well-chosen weighting function $f(x)$, the GloVe model down-weights the significance of rare word-context pairs and pays no attention to the unobserved pairs. Explicitly expressed in ``negative-sampling'', the SGNS model gains its success by assuming that randomly-chosen word-context pairs takes little or even no appearance in the corpus. Meanwhile, Levy and Goldberg \cite{sgns-fact} also point out that rare words are down-weighted in SGNS's objective shown in Equation \ref{eq:l_sgns}.

The choice of weighting function $f(x)$ neglecting the unobserved word-context pairs is for the sake of efficiency and also avoiding the appearance of undefined $\log(0)$. Whether defining an objective for the unobserved word-context pairs and taking advantage of the ``negative-sampling'' can improve the performance remains an open question.

\section{Observations of the Bias Terms in the GloVe Model}

Curious about the optimized values of the bias terms in the GloVe model and the validity of ``fixing the bias terms'' to be $\log{\#(w_i)}$ and $\log{\#(c_j)}$ in the SGNS model, we observe the trained bias terms in GloVe and compare them to the fixed term in SGNS.

We train the GloVe model on a Wikipedia dump with $1.5$ billion tokens and build a vocabulary of words occurring no less than $100$ times in the corpus. We set $x_{max}$ to be $10$ or $100$ and $\alpha$ to be $3/4$ in the weighting function $f(x)$. Word-context pairs are counted symmetrically using the same techniques given by \cite{glove}. We run $50$ iterations to train $300$-dimensional vectors for words and contexts.

\begin{figure}[!h]
	\centering
	\begin{subfigure}[b]{0.5\textwidth}
		\includegraphics[width=\textwidth]{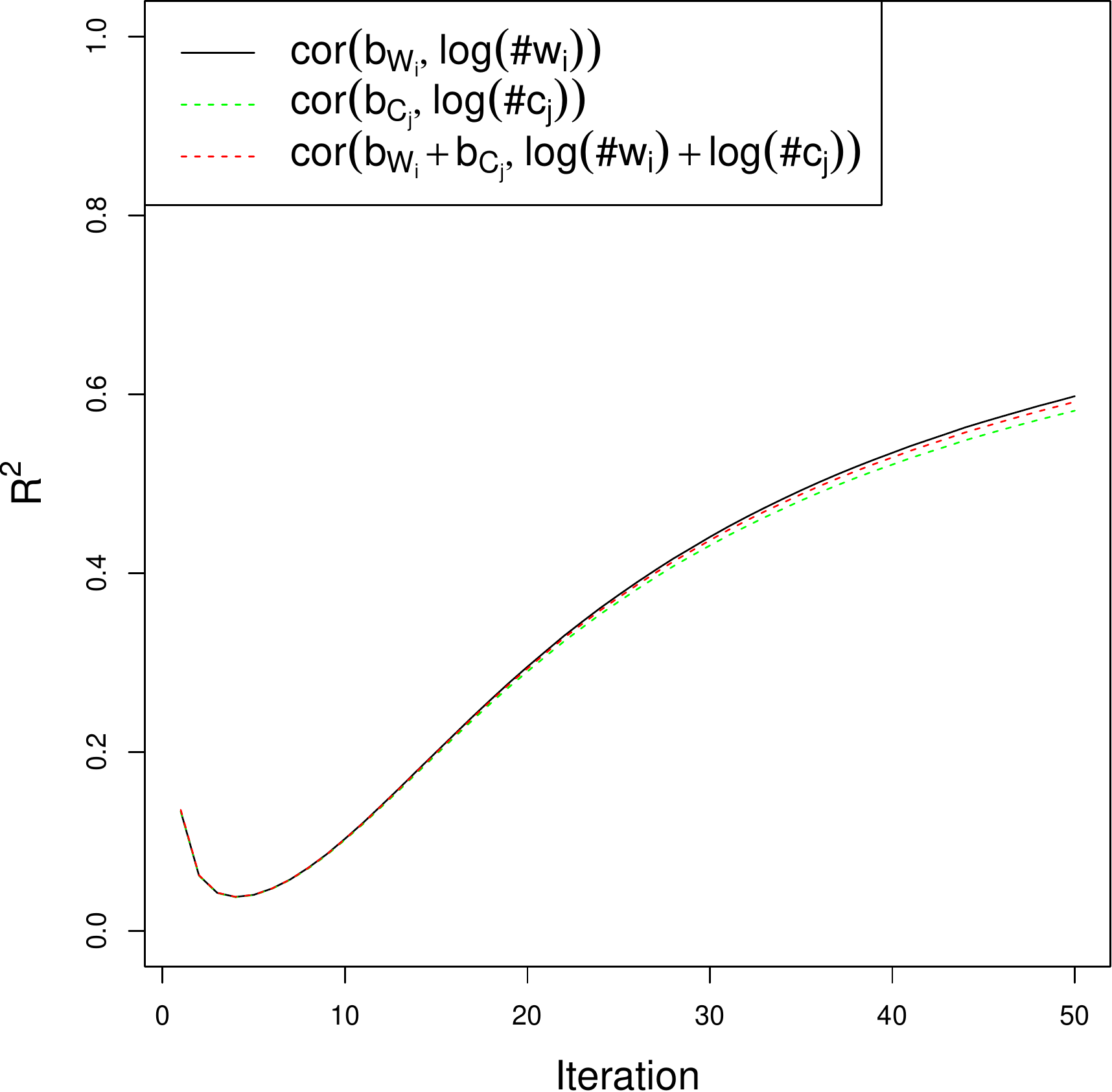}
		\caption{$x_{max}=100$}
		\label{fig:iter_100}
	\end{subfigure}%
	~
	\begin{subfigure}[b]{0.5\textwidth}
		\includegraphics[width=\textwidth]{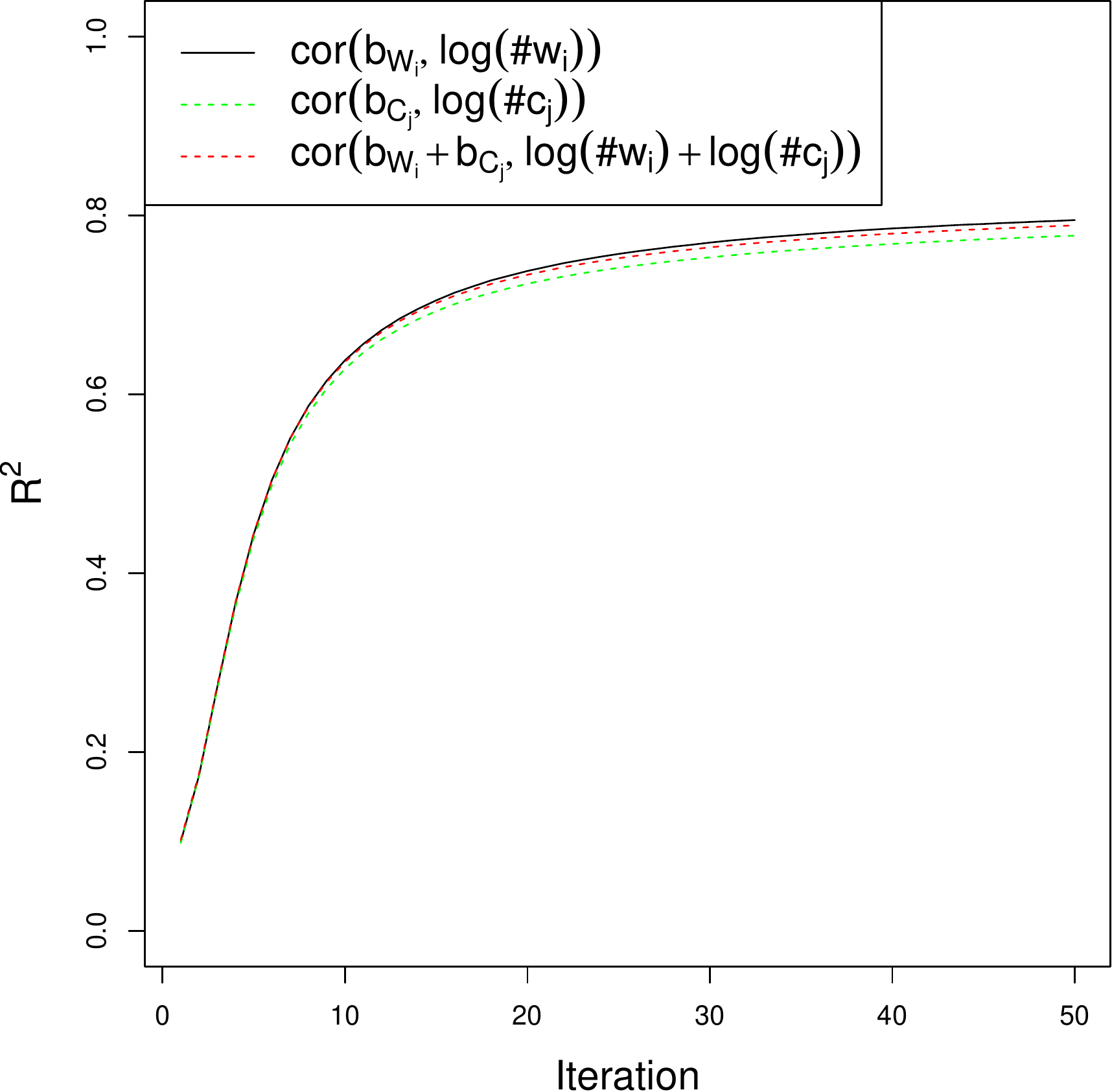}
		\caption{$x_{max}=10$}
		\label{fig:iter_10}
	\end{subfigure}
	\caption{Pearson correlation coefficient $R^2$ as a function of the number of iterations. }
	\label{fig:iterations}
\end{figure}

Figure \ref{fig:iterations} shows the Pearson correlation between $b_{W_i}$ and $\log{\#(w_i)}$, between $b_{C_j}$ and $\log{\#(c_j)}$ and between $b_{W_i} + b_{C_j}$ and $\log{\#(w_i)} + \log{\#(c_j)}$ with different $x_{max}$ values. 

\begin{figure}[!h]
	\centering
	\begin{subfigure}[b]{0.5\textwidth}
		\includegraphics[width=\textwidth]{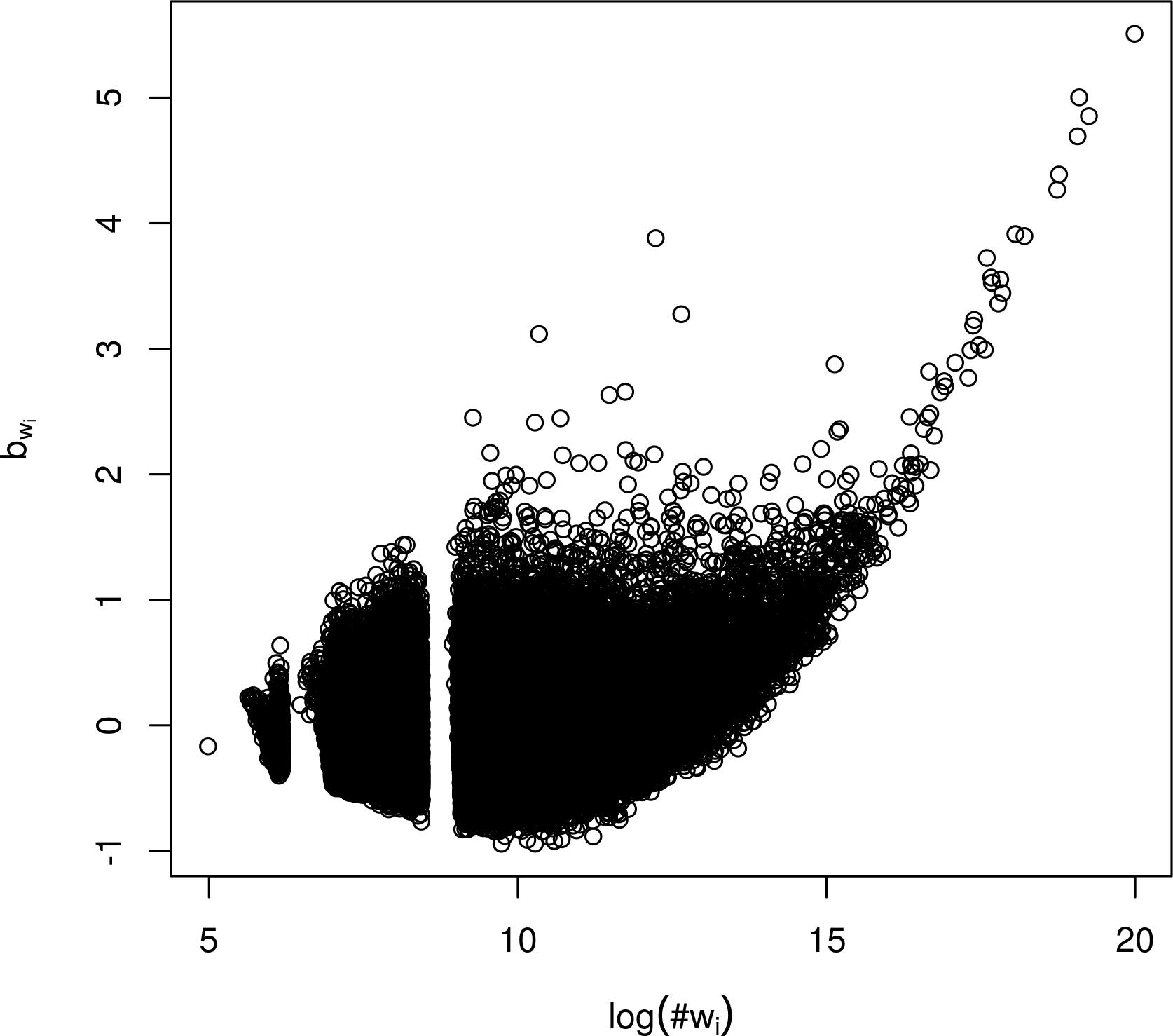}
		\caption{$x_{max}=100, iter=1, R=0.366$}
		\label{fig:100_1}
	\end{subfigure}%
	~
	\begin{subfigure}[b]{0.5\textwidth}
		\includegraphics[width=\textwidth]{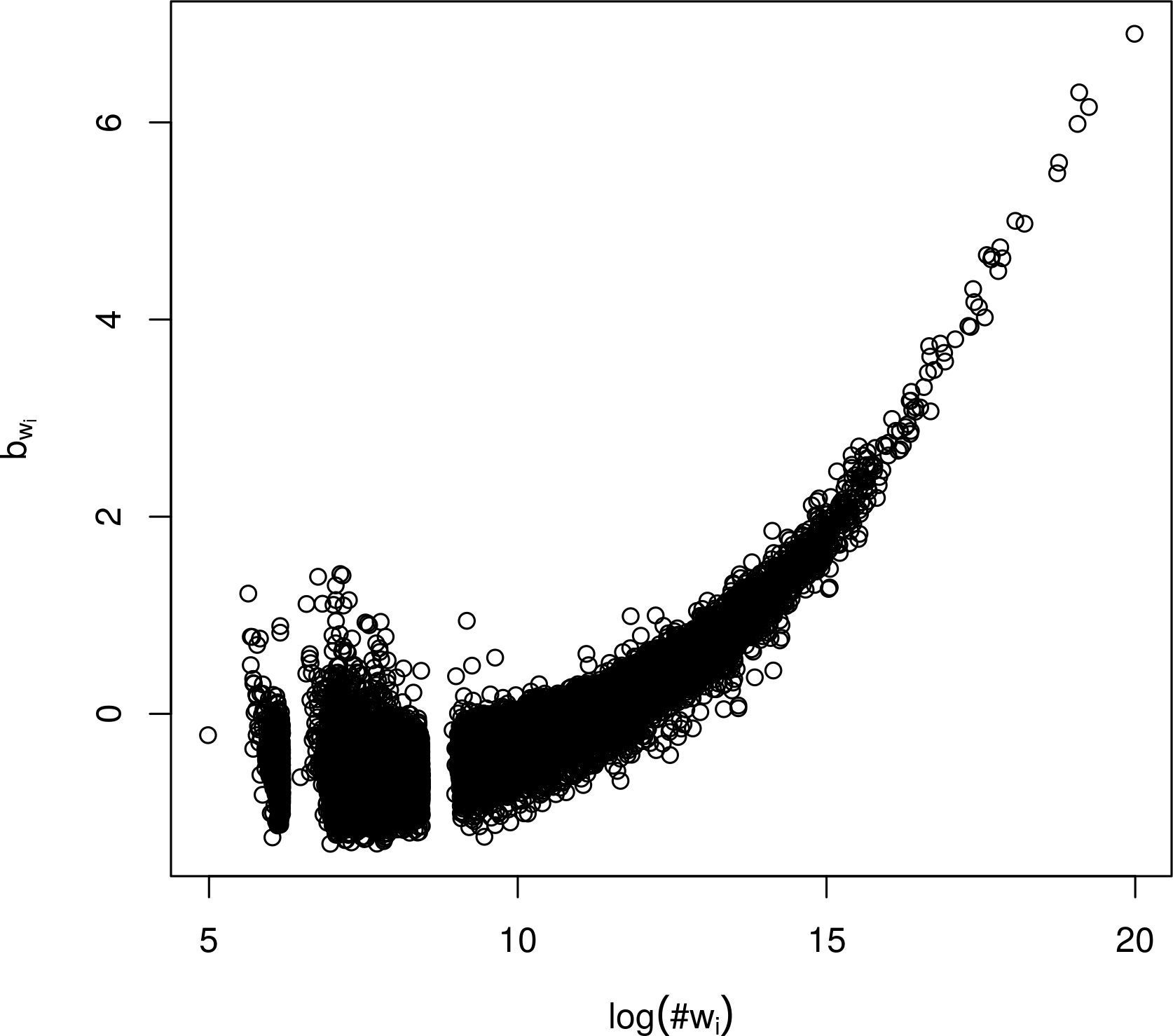}
		\caption{$x_{max}=100, iter=50, R=0.773$}
		\label{fig:100_50}
	\end{subfigure}
	
	\begin{subfigure}[b]{0.5\textwidth}
		\includegraphics[width=\textwidth]{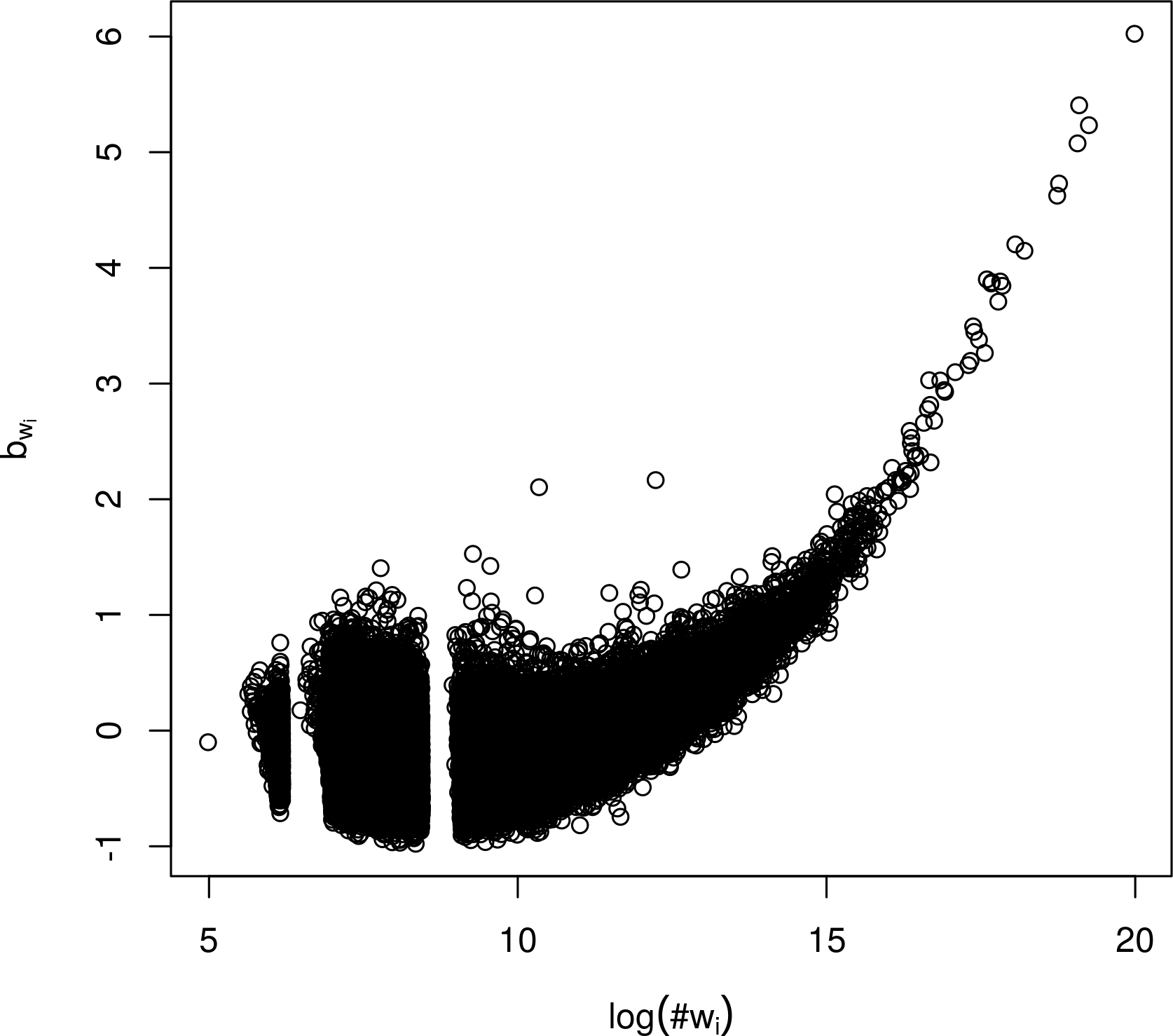}
		\caption{$x_{max}=10, iter=1, R=0.316$}
		\label{fig:10_1}
	\end{subfigure}%
	~
	\begin{subfigure}[b]{0.5\textwidth}
		\includegraphics[width=\textwidth]{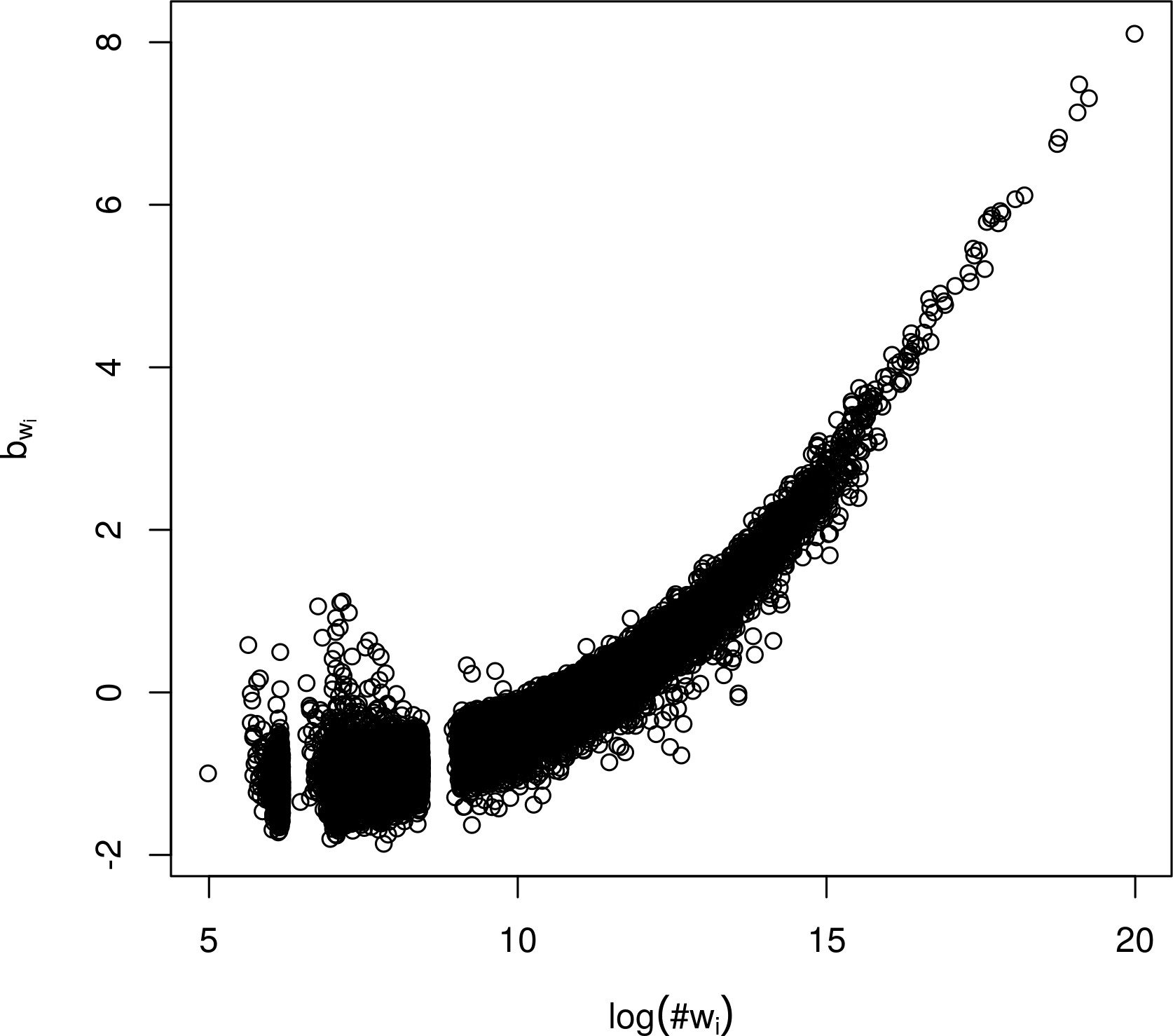}
		\caption{$x_{max}=10, iter=50, R=0.892$}
		\label{fig:10_50}
	\end{subfigure}
	\caption{Distribution of $b_{W_i}$ as a function of $\log{\#(w_i)}$
		after the first iteration and after all 50 iterations.
		Pearson correlation coefficients $R$ are given. }
	\label{fig:plot}
\end{figure}

Figure \ref{fig:plot} depicts the distribution of $b_{W_i}$ with respect to $\log{\#(w_i)}$ after the first iteration and after all 50 iterations. The two bands in the graph may be due to truncation of less frequent words.

We see that $b_{W_i}$ correlates well to $\log{\#(w_i)}$ after 50 iterations, and that less weighting effect (with smaller $x_{max}$) results in a higher correlation. Though not explicitly written in the objective function, GloVe is actually optimizing $W_{i}\cdot C_{j}^T$ towards a shifted-PMI, just like what is done in the SGNS model.

\section{Discussion}
We show that interestingly, GloVe and SGNS, one explicitly factorizing a co-occurrence matrix and one implicitly factorizing a shifted-PMI matrix, are actually sharing similar objectives, though not completely the same. The training objective of SGNS is similar to the one of a specialized form of GloVe. Their differences mainly come from different cost functions and weighting strategies. Further we observe that in empirical experiments, the bias terms in the GloVe model tend to converge toward the corresponding terms in the SGNS model. We suppose that this may be a good approximation for the globally optimized value.

Future investigation may focus on the choices of the weighting function and their effect on the two models.

\bibliographystyle{plain}
\bibliography{glove_sgns}

\end{document}